\documentclass[9pt,twocolumn,oneside]{article}
\pdfoutput=1

\usepackage{arxiv}

\usepackage[utf8]{inputenc} 
\usepackage[T1]{fontenc}    
\usepackage{hyperref}       
\usepackage{url}            
\usepackage{booktabs}       
\usepackage{amsfonts}       
\usepackage{amsmath}
\usepackage{nicefrac}       
\usepackage{microtype}      
\usepackage{lipsum}		
\usepackage{graphicx}
\usepackage{doi}
\usepackage{colortbl}
\usepackage[T1]{fontenc}
\usepackage{tabularx}
\usepackage{caption}
\usepackage{cleveref}
\usepackage{subcaption}
\usepackage{balance}

\title{Quantifying Itch and its Impact on Sleep Using Machine Learning and Radio Signals}

\author{{Michail Ouroutzoglou\textsuperscript{1}} \\
	Massachusetts Institute of Technology\\
	\texttt{michail@mit.edu} \\
	\And
	{Mingmin Zhao} \\
	Computer and Information Science Department\\
    University of Pennsylvania \\
    \texttt{mingminz@cis.upenn.edu} \\
	\And
	{Joshua Hellerstein} \\
	Massachusetts Institute of Technology\\
    \texttt{jhellerstein96@gmail.com} \\
	\And
	{Hariharan Rahul} \\
	Emerald Innovations Inc.\\
    \texttt{rahul@emeraldinno.com} \\
	\And
	{Asima Badic} \\
	Washington University School of Medicine\\
    \texttt{asimabadic@wustl.edu} \\
    \And
	{Brian S.~Kim\textsuperscript{1}} \\
	Kimberly and Eric J. Waldman Department of Dermatology\\
    Mark Lebwohl Center for Neuroinflammation and Sensation\\
    Marc and Jennifer Lipschultz Precision Immunology Institute\\
    Friedman Brain Institute, Icahn School of Medicine at Mount Sinai\\
	\texttt{itchdoctor@mountsinai.org} \\
	\And
	{Dina Katabi\textsuperscript{1}} \\
	Massachusetts Institute of Technology\\
    Emerald Innovations Inc.\\
	\texttt{dk@mit.edu} \\
}

\date{}

\renewcommand{\headeright}{}
\renewcommand{\undertitle}{}
\renewcommand{\shorttitle}{Quantifying Itch and its Impact on Sleep Using Machine Learning and Radio Signals}

\hypersetup{
pdftitle={Quantifying Itch and its Impact on Sleep Using Machine Learning and Radio Signals},
pdfsubject={cs.LG, cs.AI, cs.CY},
pdfauthor={Michail Ouroutzoglou, Mingmin Zhao, Joshua Hellerstein, Hariharan Rahul, Asima Badic, Brian S.~Kim, Dina Katabi},
pdfkeywords={Chronic Pruritus, Sleep quality, Self-Reported Assessment, Machine Learning, Wireless Sensing}}

\begin{document}

\twocolumn[\begin{@twocolumnfalse}
\maketitle

\begin{abstract}
	Chronic itch affects 13\% of the US population, is highly debilitating, and underlies many medical conditions. A major challenge in clinical care and new therapeutics development is the lack of an objective measure for quantifying itch, leading to reliance on subjective measures like patients’ self-assessment of itch severity. In this paper, we show that a home radio device paired with artificial intelligence (AI) can concurrently capture scratching and evaluate its impact on sleep quality by analyzing radio signals bouncing in the environment. The device eliminates the need for wearable sensors or skin contact, enabling monitoring of chronic itch over extended periods at home without burdening patients or interfering with their skin condition. To validate the technology, we conducted an observational clinical study of chronic pruritus patients, monitored at home for one month using both the radio device and an infrared camera. Comparing the output of the device to ground truth data from the camera demonstrates its feasibility and accuracy (ROC AUC = 0.997, sensitivity = 0.825, specificity = 0.997). The results reveal a significant correlation between scratching and low sleep quality, manifested as a reduction in sleep efficiency (R = 0.6, p < 0.001) and an increase in sleep latency (R = 0.68, p < 0.001). Our study underscores the potential of passive, long-term, at-home monitoring of chronic scratching and its sleep implications, offering a valuable tool for both clinical care of chronic itch patients and pharmaceutical clinical trials.
\end{abstract}


\end{@twocolumnfalse}]
\let\thefootnote\relax\footnotetext{\textsuperscript{1}To whom correspondence should be addressed. E-mail: michail@mit.edu, itchdoctor@mountsinai.org or dk@mit.edu}

\clearpage
\section{Introduction}
\vspace{-5pt}
Chronic pruritus affects about 13\% of the US population, resulting in an annual expenditure of over \$90 billion \cite{Matterne_2011}. It is as debilitating as chronic pain, profoundly impacting sleep and quality of life \cite{Erturk_2012, Halvorsen_2009, Kini_2011, Pisoni_2006}. A primary challenge in both itch management and drug development is the lack of an objective measure for quantifying itch \cite{Silverberg_2017}.  Currently, the clinical standard is the numerical rating scale (NRS), where patients rate the severity of their itch over the past 24 hours on a scale from 0 ("no itch") to 10 ("worst imaginable itch") \cite{Yosipovitch_2019}. Other assessment tools, like the visual analog scale (VAS) and the 5-D itch scale, exist but are similarly subjective \cite{Murray_2011, Elman_2009}. Such self-assessment methods lack sensitivity to small changes, are hard to compare across patients, and can be unreliable in children and the elderly, even though chronic itch is common in these groups \cite{Wootton_2012, Camfferman_2010}.

\begin{figure*}[h!]
\centering
\includegraphics[width=.8\linewidth]{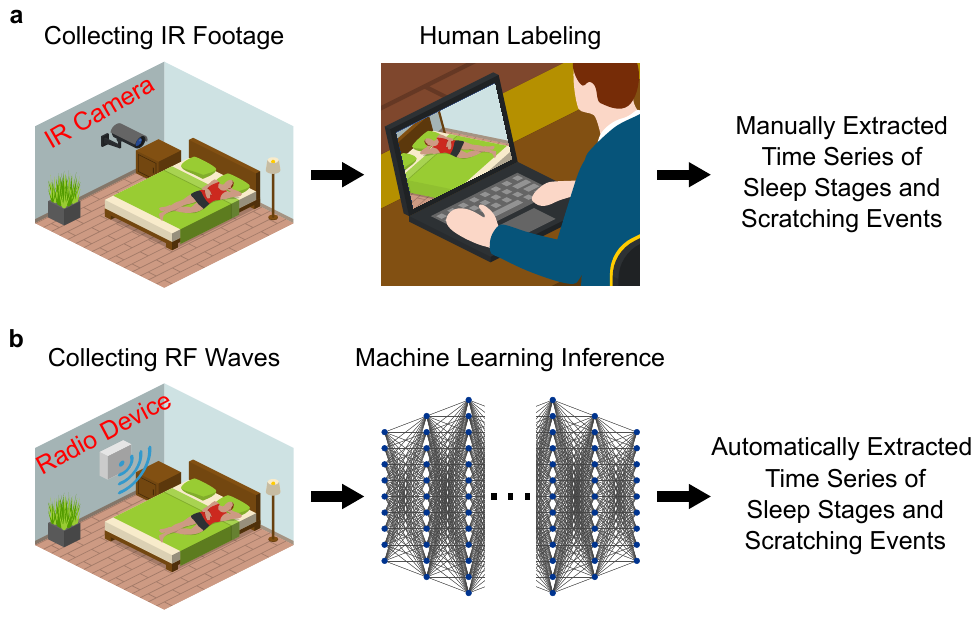}
\begin{minipage}{0.8\textwidth}
\phantomsubcaption{\label{fig:fig1a}}
\end{minipage}
\begin{minipage}{0.8\textwidth}
\phantomsubcaption{\label{fig:fig1b}}
\end{minipage}
\caption{\textbf{Illustration of video-based and radio-based assessment of scratching and sleep.} (a) Assessment of nocturnal scratching and sleep through human annotation of videos of patients in the bedrooms. (b) Assessment of nocturnal scratching and sleep by analyzing the radio frequency (RF) signals that bounce off patients during sleep using machine learning.
\label{fig:fig1}}
\end{figure*}

Nocturnal scratching has been seen as a promising objective measure of itch \cite{Elman_2009, Erickson_2019, Phan_2012, Ebata_1996, Ebata_1999, Smith_2019}. Scratching is objective, quantifiable, and has been routinely measured in animal studies to assess itch \cite{Shimada_2008}. It has also been shown sensitive across different types of itch including inflammatory, neuropathic, and even dry skin itch \cite{Liu_2009, Zhao_2013, Komaromy_2018, Wang_2021, Trier_2022}. Yet, monitoring nocturnal scratching in humans is challenging. The straightforward approach involves installing infrared (IR) cameras in patients' bedrooms. Human observers then review the videos and annotate scratching occurrences, as shown in ~\cref{fig:fig1a}. This method raises privacy concerns and incurs excessive overhead and costs due to the extensive volume of video footage that needs manual review. Additionally, clinical trials for chronic itch require daily assessments over several months \cite{dupilumab_pn_2019, dupilumab_cpuo_2022, dupilumab_2020}, exacerbating privacy and cost concerns.  Automating this task with computer vision is not straightforward either, as much scratching occurs under blankets and needs scrutiny and consensus from multiple human labelers. Consequently, while video recording is recognized as the ground truth for scratching, it is deemed impractical for extensive use in clinical trials or care.

The challenges with video monitoring have led researchers to explore wearable devices, resulting in hundreds of papers on the topic \cite{Chun_2021, Yang_2023, Moreau_2018, Mahadevan_2021, Ji_2023, Ikoma_2019, Petersen_2013, Benjamin_2004, Ebata_2001, Lee_2015}.
Table S1 in the supplementary material summarizes this literature.
The most prevalent method employs wrist actigraphy to monitor scratching.
The performance of wrist actigraphy however is relatively low \cite{Moreau_2018, Ji_2023, Petersen_2013}, which has led some researchers to propose soft sensors mounted on the back of the hand with adhesive support to offer better accuracy \cite{Chun_2021, Yang_2023}. However, removal and re-adhesion of such sensors increases the risk of skin irritation, particularly in clinical trials which typically last several months \cite{dupilumab_pn_2019, dupilumab_cpuo_2022, dupilumab_2020}. Additionally, wearable devices detect scratching only if performed with the hand wearing the device, but it is common for patients to scratch with other body parts or rub against bed sheets.

Thus, despite significant advancements, there remains a need for an objective, accurate scratching measure that is not limited to hand movements and can monitor scratching for months without the potential of interfering with skin condition. Ideally, this solution can concurrently evaluate scratching and its impact on sleep. Impaired sleep drastically affects itch patients' quality of life and has been linked to neurocognitive deficits in children with chronic eczema \cite{Camfferman_2010} and cardiovascular death in patient with uremic pruritus \cite{Weng_2018}.

Here, we introduce a new approach that simultaneously measures scratching and sleep in an objective, sensitive, privacy-preserving, and contactless manner. We build on, and expand, recent literature that demonstrated the feasibility of using radio frequency (RF) signals to passively measure physiological signals (e.g., motion, vital signs, and sleep) in Parkinson’s, Alzheimer’s, COVID-19, and others \cite{Kabelac_2019, Vahia_2020, Loring_2019, Liu_2022, Yang_2022, Zhang_2021, Zhao_2021}. Our approach uses a radio device that sits in the background of the home like a Wi-Fi router. It sends out wireless signals (1000 times lower power than home Wi-Fi) and collects their reflections from the environment and nearby people \cite{Kabelac_2019, Vahia_2020, Loring_2019, Adib_2015, Adib_2014, Adib_2013, Adib_2015b, Adib_2015c, Hsu_2017, Li_2019, Tian_2018, Yue_2018, Zhao_2016, Zhao_2018, Zhao_2018_b, Zhao_2017, Hsu_2017b, Kabelac_2020}. We zoom in on radio signals reflected from the participant’s bed using beamforming techniques \cite{Adib_2015c, Hsu_2017b}. We then analyze these radio signals using custom neural networks to infer nocturnal scratching and sleep. To infer scratching instances, we develop a novel neural network model. The network first identifies repetitive motion patterns that characterize scratching, then processes those patterns to output a time series labeled with the occurrence of scratching. To infer sleep, we leverage a pre-existing AI model for sleep monitoring using radio signals \cite{Zhao_2017, Hsu_2017b}. Crucially, our method is unobtrusive, eliminating the need for patients to wear sensors or self-report symptom severity. \Cref{fig:fig1b} illustrates our solution.

To validate our solution, we conducted an observational clinical study, in which we monitored 20 chronic itch patients for one month in their homes. Benchmarking our device's output against infrared video recordings – serving as the ground truth – revealed that the device is highly accurate in detecting scratching (ROC AUC = 0.997, sensitivity = 0.825, specificity = 0.997), and correlating it with reduction in sleep quality (R = 0.6, p < 0.001 for sleep efficiency; R = 0.68, p < 0.001 for WASO; and R = 0.6, p < 0.001 for sleep latency). Interestingly, the device can track scratching even when zooming in on individual patients, which is necessary for clinical care (0.71 < R < 0.99 with p < 0.003 for n=1).  Our findings have also confirmed previous studies, showing that subjective self-assessments have a weak correlation with objective video-based scratching evaluations (R = 0.22 p < 0.001).

Our research underscores the potential of integrating a Wi-Fi-like radio device with machine learning to objectively and passively monitor nocturnal scratching and its impact on sleep. This new clinical tool promises the ability to test responses to treatments in both clinical practice and interventional clinical trials.


\section{Results}
\subsection{Study Participants}
Twenty adults diagnosed with chronic itch (with an Investigator’s Global Assessment (IGA) mean of 2.8 and a standard deviation of 1.5) were recruited for a prospective observational study and provided informed consent under IRB 201811086. All participants reported itching for a minimum of 6 weeks prior to enrollement. The study aimed to validate the efficacy of utilizing radio signals for scratching detection and evaluate the correlation between scratching and sleep parameters. To ensure the evaluation of scratching detection's accuracy was independent of the specific itch condition, we selected participants spanning a range of pruritus diagnoses, including atopic dermatitis (AD), prurigo nodularis (PN), and chronic pruritus of unknown origin (CPUO). Demographics and baseline disease characteristics are shown in Table 1. 

\begin{table}[h]
\centering

\begingroup
\setlength{\tabcolsep}{10pt} 
\renewcommand{\arraystretch}{1.5}
\begin{tabular}{l c} 
 \arrayrulecolor{black}\hline
 \multicolumn{2}{c}{\textbf{Characteristic Pruritus}, N = $20$} \\
 \hline 
 \textbf{Age}$^1$ & 57.5 (16.1) \\ 
 \hline
 \textbf{Gender}$^2$ &  \\ 
 \hline
 ~~~~Female & 14 (70\%) \\
 \hline
 ~~~~Male & 6 (30\%) \\
 \hline
 \textbf{Race}$^2$ &  \\ 
 \hline
 ~~~~Black & 4 (20\%) \\
 \hline
 ~~~~White & 16 (80\%) \\
 \hline
 \textbf{Diagnosis}$^2$ &  \\ 
 \hline
 ~~~~AD & 4 (20\%) \\
 \hline
 ~~~~PN & 2 (10\%) \\
 \hline
 ~~~~CPUO & 14 (70\%) \\
 \hline 
 \textbf{NRS}$^1$ & 7.9 (2.0) \\ 
 \hline 
 \textbf{IGA}$^1$ & 2.8 (1.5) \\ 
 \hline
 \multicolumn{2}{l}{$~^1$Mean (SD);$~^2$n (\%)} \\
\end{tabular}
 \caption{Characteristics of the population.}
 \label{table:1}
 
\endgroup
\end{table}

The participants were monitored at home for 4 weeks using both the radio device and an infrared camera installed in the bedroom. Participants were asked to report their daily NRS and sleep scores (on a scale of 0 to 10) for at least 4 nights per week. We analyzed nights for which there is data from both the infrared camera and the radio device, excluding nights shorter than two hours. In total, we analyzed 364 nights of data, 2721 hours, and 21,619 scratching bouts (i.e., scratching events). To our knowledge, this constitutes the largest scratching dataset with ground truth labels. 

To evaluate the accuracy of the radio device, we compared its output with the ground truth scratching from manual annotations of video recordings. Each video was labeled by three labelers, on average, and a majority vote was performed to obtain the ground truth labels. We employed a 4-fold cross-validation method for training and testing our model. In this approach, the data is segmented into four subsets. The neural network is then trained on three of these subsets and tested on the fourth. This process is replicated four times, each with different test set. Importantly, we ensured that a participant never simultaneously appeared in both the training and testing subsets.

For sleep assessment, we used a pre-existing machine learning model developed by Zhao et al \cite{Zhao_2017}. Similar to standard sleep lab measurements, this model produces a sleep hypnogram, i.e., a time series where each 30-second epoch is labeled with the participant’s sleep stage: wake, light sleep, deep sleep, or REM. It achieves a classification accuracy of 79.8\% per 30-second sleep epoch \cite{Zhao_2017}.

\subsection{Performance of Nocturnal Scratching Detection}
\begin{figure*}[p!]
\centering
\includegraphics[width=.8\linewidth]{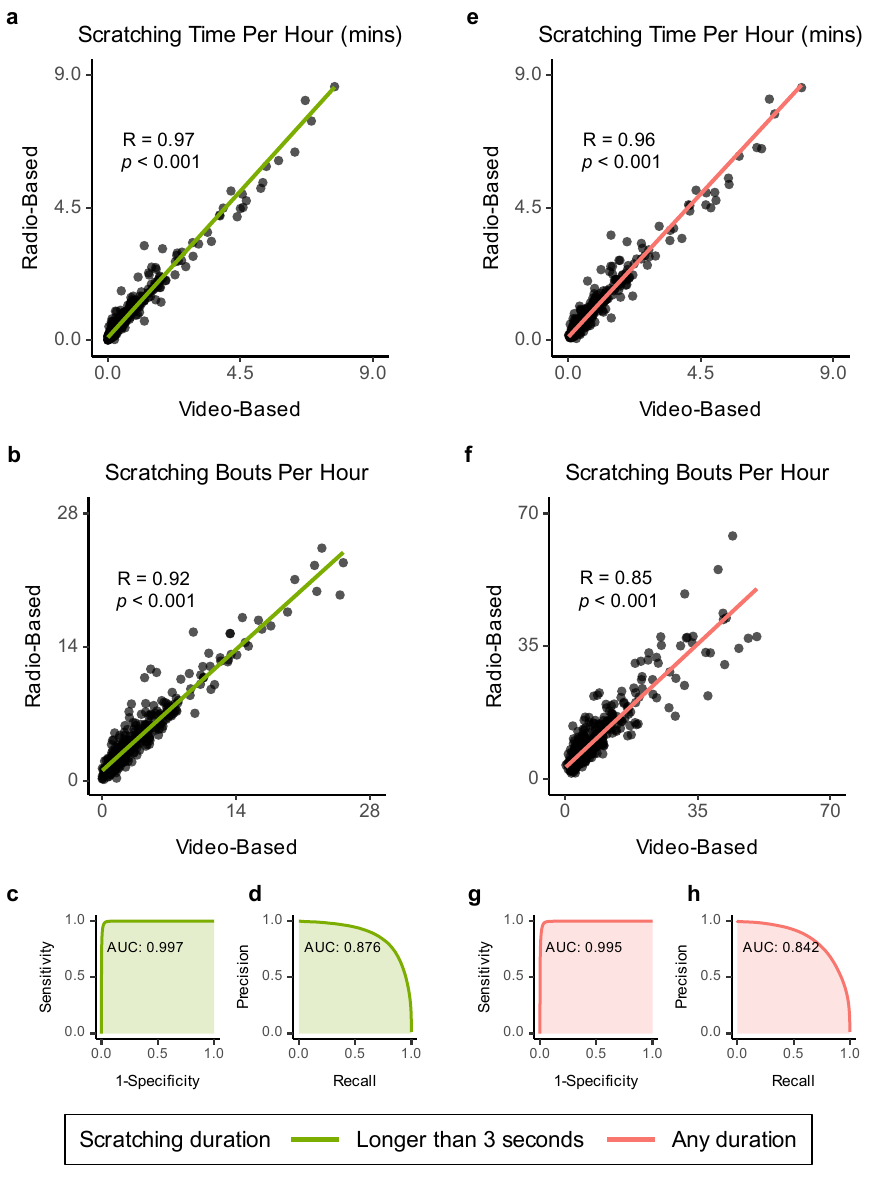}
\begin{minipage}{0.4\textwidth}
\phantomsubcaption{\label{fig:fig2a}}
\phantomsubcaption{\label{fig:fig2b}}
\phantomsubcaption{\label{fig:fig2c}}
\phantomsubcaption{\label{fig:fig2d}}
\end{minipage}
\begin{minipage}{0.4\textwidth}
\phantomsubcaption{\label{fig:fig2e}}
\phantomsubcaption{\label{fig:fig2f}}
\phantomsubcaption{\label{fig:fig2g}}
\phantomsubcaption{\label{fig:fig2h}}
\end{minipage}
\caption{\textbf{Performance of radio-based scratching assessment.} The left column presents results for scratching lasting longer than 3 seconds, while the right column presents results for all scratching occurrences. (a), (b) compare the scratching time per hour (STH) and scratching bouts per hour (SBH) estimated by our radio-based approach to manual annotations of scratching in videos, for scratching lasting more than 3 seconds. Each point in the figures corresponds to a full night of data for one patient (totaling 364 nights across n=20 participants). The figures show that the assessment of scratching using the radio device has a strong correlation with the camera-based ground truth (R = 0.95, p < 0.001 for STH, R = 0.83, p < 0.001 for SBH). (c) shows the ROC curve and the AUC of radio-based scratching assessment (ROC AUC = 0.997). (d) shows the Precision-Recall curve and the PR AUC of radio-based scratching assessment (PR AUC = 0.876). (e)-(h) plot the STH correlation, SBH correlation, ROC curve, and Precision-Recall curve, respectively, comparing the scratching predicted by the radio device against the ground truth from video recording, for all scratching occurrences regardless of their duration.}
\label{fig:fig2}
\end{figure*}
To evaluate the performance on scratching detection, we consider two settings. First, it has been customary in the literature to focus on scratching events (continuous scratching movements) lasting 3 seconds or longer \cite{Chun_2021, Yang_2023, Mahadevan_2021, Ji_2023, Ikoma_2019}, as such longer events are considered more indicative of disease severity. Taking a similar approach, we filtered the ground truth data and the model predictions to exclude scratching events shorter than 3 seconds. Then, for each time instance (time is sampled at 15 Hz), we compare the model’s prediction with the ground truth data from video recordings. The results demonstrate that the model offers high performance:

\begin{figure*}[p!]
\centering
\includegraphics[width=1.\linewidth]{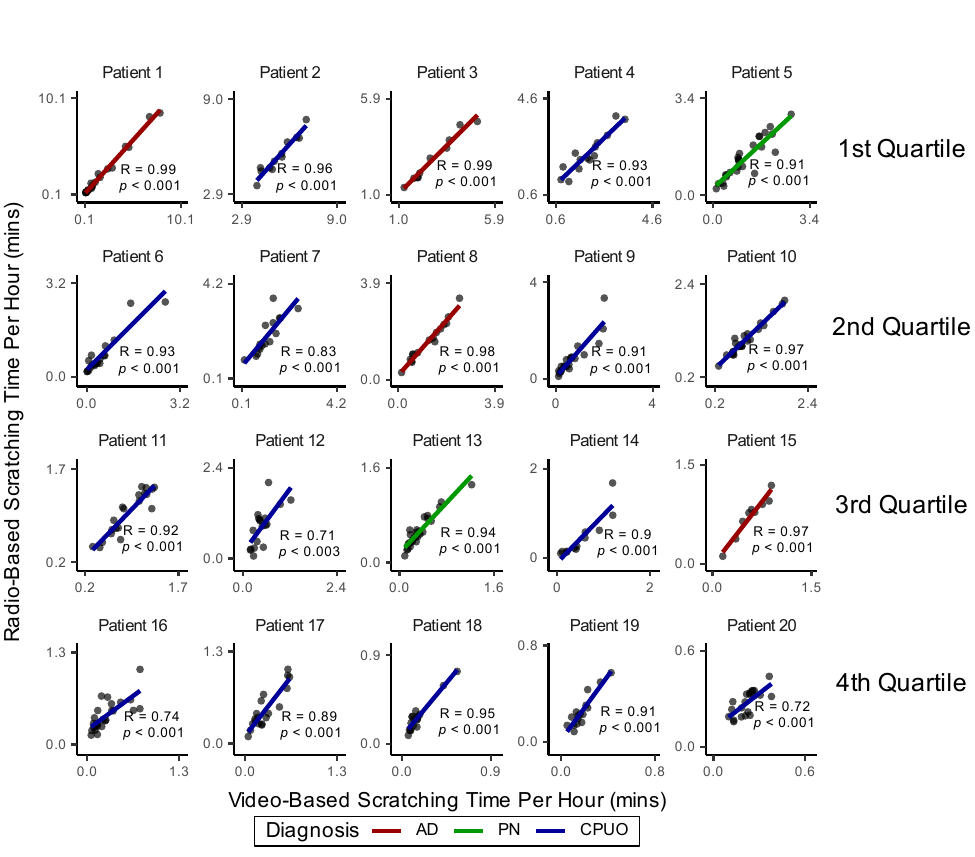}
\caption{\textbf{Per-participant performance evaluation of radio-based scratching assessment.} Each plot shows the Pearson’s correlation between the scratching time per hour inferred from radio signals and the ground truth video annotations for a particular participant. Each point in these plots corresponds to a full night of data for the participant. The participants are ordered with respect to their maximum scratching time per hour across all the nights, from highest to lowest. The scratching time measured by the radio device is significantly correlated with the ground truth scratching time per hour for all participants across different disease diagnoses (color coded with red, green, and blue for AD, PN, and CPUO, respectively) and varying disease severity. }
\label{fig:fig3}
\end{figure*}

\begin{itemize}
    \item The scratching time per hour (STH), i.e., the amount of nocturnal scratching divided by the length of the night in hours, displays a strong correlation between the model’s output and the ground truth (R = 0.97, p < 0.001) (\cref{fig:fig2a}). This analysis uses the repeated measures correlation (rmcorr) \cite{Bakdash_2017} to account for differences in the numbers of nights from different participants. 
    \item Similarly, rmcorr showed that the scratching bouts per hour (SBH), i.e., the number of nocturnal scratching events divided by the length of the night in hours, exhibits a strong correlation between the model and the ground truth (R = 0.92, p < 0.001) (\cref{fig:fig2b}). 
    \item The ROC curve analysis shows that the model delivers an AUC of 0.997, and achieves a specificity of 0.997 (95\% confidence interval (CI) [0.997, 0.997]), and sensitivity of 0.825 (95\% CI [0.824, 0.826]) (\cref{fig:fig2c}). 
    \item The Precision-Recall AUC is 0.876, and the model operates at a precision of 0.766 (95\% CI [0.765, 0.767]), a recall of 0.825, and F1-score of 0.795 (\cref{fig:fig2d}).
\end{itemize}

Next, we expanded the scope to consider all scratching events including those lasting less than 3 seconds. Naturally, it is harder to detect shorter events, yet the model continues to deliver high performance with statistical significance, as shown in ~\cref{fig:fig2e,fig:fig2f,fig:fig2g,fig:fig2h}. Specifically, both the scratching time per hour and the scratching bouts per hour are strongly correlated between the model’s output and the ground truth (R = 0.96, p < 0.001 for STH, and R = 0.85, p < 0.001 for SBH).  The ROC AUC is 0.995 and the specificity is 0.995 and the sensitivity is 0.806. The Precision-Recall AUC is 0.842, and the model operates at a precision of 0.707, a recall of 0.806, and F1-score of 0.753.

These findings collectively affirm the AI model's capacity to identify scratching instances, irrespective of whether the focus is on extended scratching bouts (longer than 3 seconds) or all-encompassing scratching events. For the rest of the results in the paper, we will consider all scratching events. 

\subsection{Performance on Scratching Assessment for N=1}
While our earlier results focus on cohort performance—crucial for clinical trials—we subsequently shifted our attention to individual-level (n=1) performance. This granular analysis is vital in clinical care, enabling the physician to track the disease burden and the medication response in a particular patient.

To delve deeper into this aspect, we plotted the data for each participant separately in \cref{fig:fig3}. The radio-based STH predictions (displayed on the y-axis) were compared with the ground truth video-based annotations of STH (shown on the x-axis), spanning all recorded nights. For clarity, participants were grouped and organized into quartiles, according to the severity of their scratching (measured by peak STH), descending from the highest to the lowest. Additionally, various chronic pruritus diagnoses were depicted using distinct colors: atopic dermatitis (AD) in red, prurigo nodularis (PN) in green, and chronic pruritus of unknown origin (CPUO) in blue.

The correlation between the model’s predictions and the ground truth assessments of scratching continues to be significant, even on a per-participant basis. Pearson’s correlation ranged from 0.71 to 0.99 across different participants, with a p-value less than 0.003 consistently across the board.

\subsection{Relationship between Scratching and Sleep}

\begin{figure*}[h!]
\centering
\includegraphics[width=.8\linewidth]{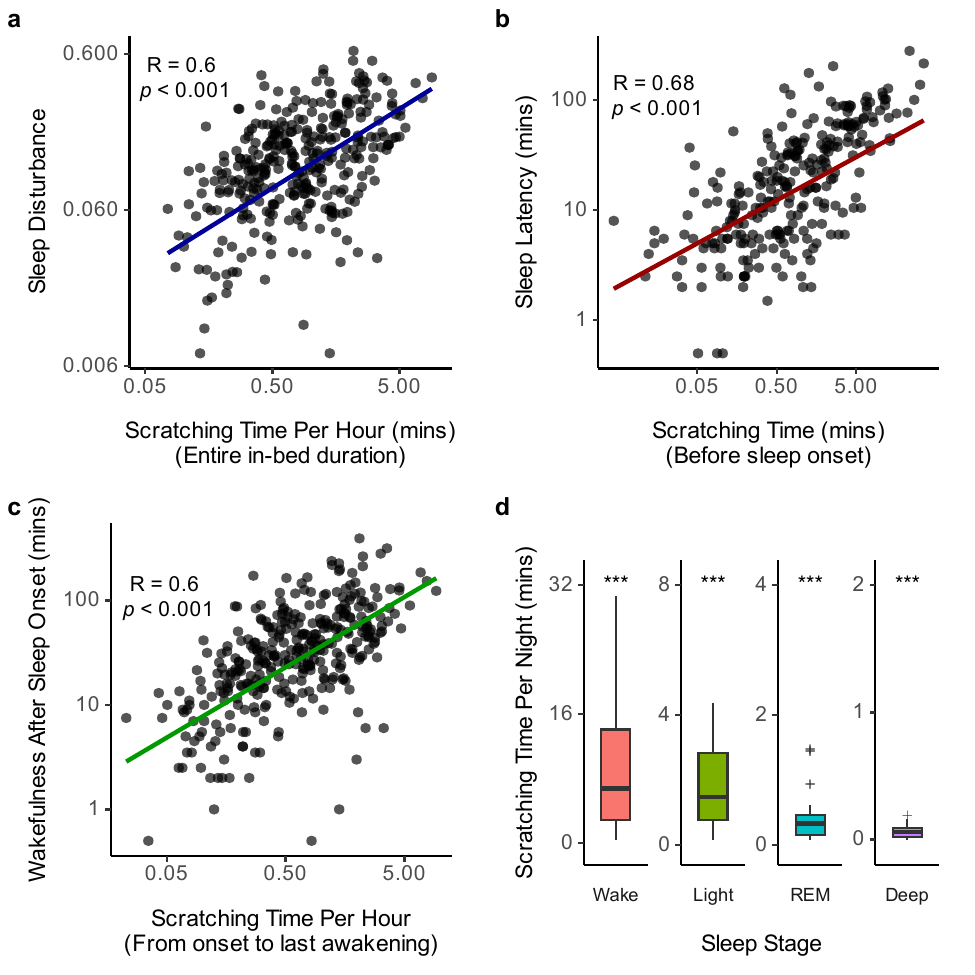}
\begin{minipage}{0.8\textwidth}
\phantomsubcaption{\label{fig:fig4a}}
\phantomsubcaption{\label{fig:fig4b}}
\end{minipage}
\begin{minipage}{0.8\textwidth}
\phantomsubcaption{\label{fig:fig4c}}
\phantomsubcaption{\label{fig:fig4d}}
\end{minipage}
\caption{\textbf{Relationship of scratching with sleep quality in individuals with chronic pruritus.} (a) plots the relationship between sleep disturbance and scratching, where sleep disturbance is the percentage of time spent awake out of the total time in bed (for 341 nights from 20 participants). (b) plots the relationship between sleep latency and scratching, where sleep latency is the time until sleep onset. (c) plots the relationship between WASO and scratching, where WASO is defined as the time spent awake after sleep onset. (d) shows the average scratching per night per participant for each sleep stage. One-sample one-sided Wilcoxon rank sum tests were conducted to assess significance.  In each box, the central line indicates the median, and the bottom and top edges of the box indicate the 25th and 75th percentiles, respectively. The whiskers extend to 1.5 times the interquartile range. Three asterisks indicate p < 0.001.}
\label{fig:fig4}
\end{figure*}
It is widely appreciated that severe itch negatively impacts sleep and quality of life \cite{Beck_2020, Bawany_2021}. However, an inherent paradox in patient reported outcome (PRO) measures is that sleep is an unconscious state, but PRO assessments attempt to infer sleep quality via recall in the conscious state. The radio device is uniquely well-suited to address this gap and deliver an objective assessment of the relationship between itch and sleep. Previous studies have demonstrated that the radio device used in this study delivers high quality data with regard to sleep efficiency, latency, and wakefulness \cite{Hsu_2017b}. Further, it accurately detects various stages of sleep – wake, light sleep, REM, and deep sleep – and performs these measurements at home \cite{Zhao_2017}. This presents an opportunity to concurrently measure and correlate scratching and sleep.

\begin{figure*}[h!]
\centering
\includegraphics[width=.8\linewidth]{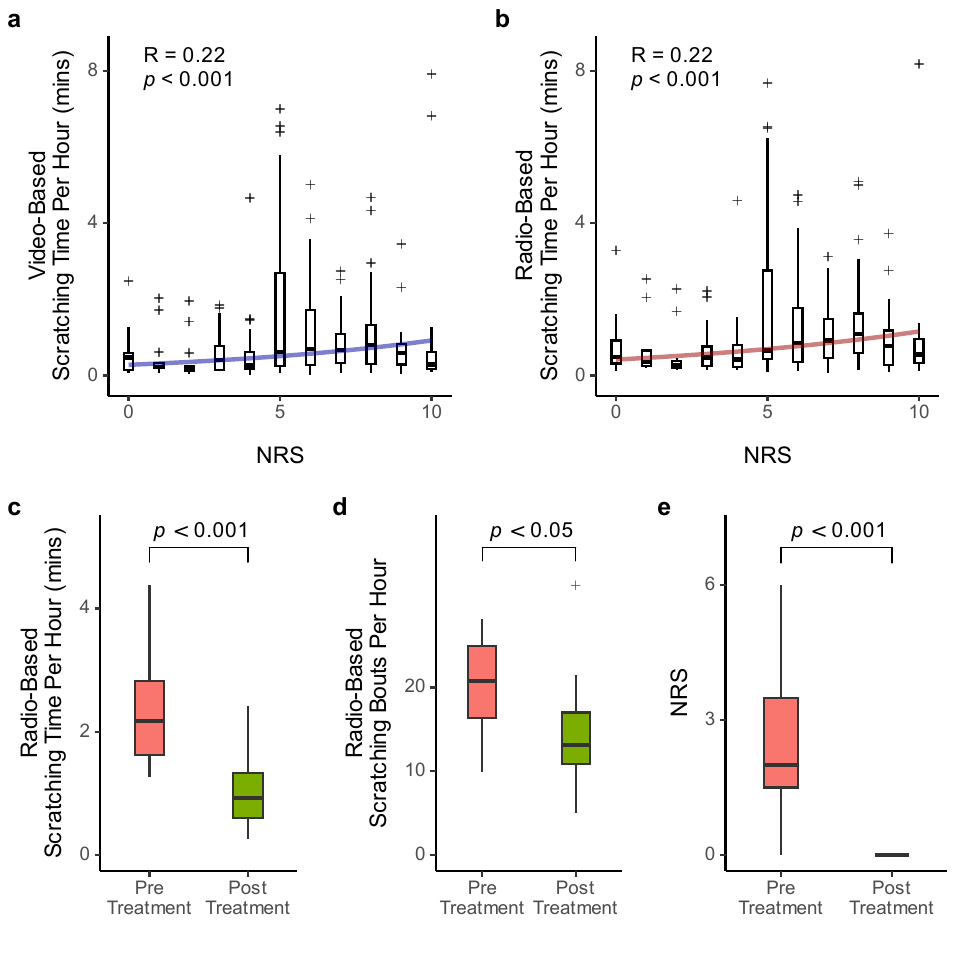}
\begin{minipage}{0.8\textwidth}
\phantomsubcaption{\label{fig:fig5a}}
\phantomsubcaption{\label{fig:fig5b}}
\end{minipage}
\begin{minipage}{0.8\textwidth}
\phantomsubcaption{\label{fig:fig5c}}
\phantomsubcaption{\label{fig:fig5d}}
\phantomsubcaption{\label{fig:fig5e}}
\end{minipage}
\caption{\textbf{Relationship between scratching measurements and self-reported NRS.} (a), (b) show scatterplots of the scratching time per hour (STH) and the self-reported NRS for each participant (342 nights from 20 participants), where scratching time is measured from video annotation in (a) and radio signals in (b). (c), (d), (e) focus on a participant that received a shot of Dupilumab 300mg during the study. In each subfigure we compare the STH, SBH, or NRS before and after one dose of treatment. Two-sample, one-sided Wilcoxon rank sum tests are performed to assess significance. Significant reduction is observed for all three cases (p < 0.001, p < 0.05 and p < 0.001, respectively). In each box plot, the central line indicates the median, and the bottom and top edges of the box indicate the 25th and 75th percentiles, respectively. The whiskers extend to 1.5 times the interquartile range. }
\label{fig:fig5}
\end{figure*}

We thus used the same wireless signals collected by the device to infer both sleep quality (using the model by Zhao et al \cite{Zhao_2017}) and scratching (using the model developed herein). Three key sleep metrics were evaluated: a) sleep disturbance, which is defined as one minus sleep efficiency, where sleep efficiency is the percentage of time the person is asleep out of the total time the person is in bed trying to fall asleep, b) sleep latency, which is the time spent awake before sleep onset, and c) wakefulness after sleep onset (WASO), which is the amount of awake time after falling asleep.  The correlation between these sleep metrics and scratching was computed using rmcorr, and the analysis was performed in log scale since the log-normal distribution provided a closer fit to the data. Nights for which the logarithmic scale is undefined were excluded. The results showed that sleep disturbance, sleep latency, and WASO were all significantly correlated with scratching (\cref{fig:fig4}). Overall, the more a participant scratched in a particular night, the higher their sleep disturbance (\cref{fig:fig4a}; R = 0.60, p < 0.001), the longer their sleep latency (\cref{fig:fig4b}; R = 0.68, p < 0.001), and the more awake time they had at night (\cref{fig:fig4c}; R = 0.60, p < 0.001). Taken together, these findings indicate that scratching imposes a high burden on sleep quality. 

Although scratching can be consciously triggered, it is often underappreciated that the itch-scratch cycle is an unconscious reflex. To examine how much scratching occurs in the unconscious state, we evaluated the amount of scratching during different stages of sleep: wake, light sleep, deep sleep, and REM (\cref{fig:fig4d}). The results demonstrate that, while most scratching occurs in the wake stage, scratching activity also occurs during sleep, even during stages of relative paralysis (i.e., REM with $p < 0.001$). These findings are consistent with prior work, which has reported that scratching could occur in all sleep stages \cite{SAVIN_1975}.

\subsection{Objective Scratching vs. Subjective Itching}
We next correlated nocturnal scratching measurements from the infrared camera and the radio device (AI) with the NRS itch score reported by the patients at bedtime (342 scores from 20 participants). We used a linear mixed-effect model (where the observed variable is log-transformed due to its log-normal distribution). Across the population, we found significant but weak correlation between the subjective NRS itch score and objective video-based (R = 0.22, p < 0.001) and radio-based measurements (R = 0.22, p < 0.001) (\cref{fig:fig5a,fig:fig5b}). Such weak correlation is consistent with prior reports \cite{Murray_2011}. Further, the weak relationship between objective ground-truth scratching from videos and patients’ self-reporting of the severity of their condition highlights the limitation of current assessment methods.

We then hypothesized that, the change in objective scratching behavior and NRS itch score during treatment and resolution of itch may demonstrate better concordance. We followed one participant with Atopic Dermatitis as they underwent treatment with Dupilumab, a monoclonal antibody that has reported broad anti-itch properties across AD, PN, and CPUO \cite{dupilumab_2020, dupilumab_cpuo_2022, dupilumab_pn_2019}. The participant received the first dose of Dupilumab (300 mg) immediately before the beginning of the monitoring period. We compared the participant’s STH, SBH, and NRS measurements two weeks before and after the administration of the second dose of Dupilumab. Indeed, there were significant reductions in both STH (\cref{fig:fig5c}; p < 0.001) and SBH (\cref{fig:fig5d}; p < 0.05). Similarly, there is also a significant reduction in self-reported NRS itch score (\cref{fig:fig5e}; p < 0.001). These results suggest that both the device and NRS have utility and overall concordance during treatment. However, even in this case, NRS can suffer from a placebo effect. For example, after receiving treatment, the patient immediately reduced their reported NRS to 0, yet the video recordings showed that the change in scratching is more gradual. Similarly, the reduction in scratching reported by the radio device was gradual and consistent with the video recordings.

\section{Discussion}
Itch is defined as an unpleasant sensation that elicits the desire or reflex to scratch. As it is only perceived by the host, itch is inherently subjective. However, scratching behavior can be observed by others and measured objectively. Scratching activity is routinely quantified in animal studies \cite{Shimada_2008}. Yet, most assessments of itch in humans have relied solely on subjective Patient-Reported Outcomes (PROs) to evaluate disease burden and drug efficacy. The absence of objective assessments of itch is often seen as a major limitation in the itch field, contributing to the high unmet need and few FDA-approved treatments for chronic pruritus. Our study shows that nocturnal scratching can be accurately monitored in humans and linked to objective impairment of sleep. Moreover, this monitoring can persist for extended periods in a completely passive manner, without burdening patients or making contact with their skin.

There have been several attempts to use wearable devices to track nocturnal scratching \cite{Murray_2011, Mahadevan_2021, Bringhurst_2004, Chun_2021, Bender_2003, Ikoma_2019, Fujita_2014, Kogure_2018, Gustafson_2014, Feuerstein_2011, Fishbein_2023, Kalinska_Bienias_2019}. These technologies offer valuable insights. However, wrist actigraphy demonstrates relatively subpar performance, and soft sensors affixed with adhesive heighten the risk of skin irritation, especially when used in several-month-long clinical trials. Thus, we believe there is a continuing need for a fully passive off-body technology that can monitor nocturnal scratching over extended periods without requiring patients to wear sensors on already irritated skin.

Further, our approach can simultaneously and passively monitor scratching and sleep. Sleep is one of the most adversely affected aspect in patients with chronic pruritus, and sleep loss significantly degrades quality of life \cite{Lee_2021}. The vast majority of studies on both sleep and itch base their conclusions on PROs instead of objective measurements. The technology we present here offers an avenue for monitoring both metrics concurrently and longitudinally in the patient’s home. Our findings indicate that increased scratching is correlated with poor sleep quality (e.g., longer sleep latency and more frequent awakenings) and that scratching occurs in all sleep stages, including REM and deep sleep.

Our results also display a mild correlation between objective nocturnal scratching and subjective PROs. Coupled with the observation that scratching can happen even during the deepest sleep stages, this points to a potential shortcoming of PROs. Specifically, they depend on the conscious recall of experiences that occur unconsciously, such as nocturnal itch and sleep.

Our method offers a beneficial tool for the clinical care of chronic pruritus patients, allowing clinicians to observe a patient's response to therapy. This can guide clinicians in making prompt decisions, crucial for drugs like corticosteroids where the risk of adverse effects escalates with extended use \cite{Stacey_2021}.  The monitoring may also aid physicians in differential diagnosis. Itch is sometimes stigmatized in clinical settings, occasionally being misconstrued as a psychiatric disorder. For instance, delusional parasitosis is a psychiatric condition where individuals believe that insects are emerging from their skin. Owing to a strong urge to touch the skin, this condition often manifests as scratching. Yet, unlike itch, such compulsive or delusional behaviors are unlikely during sleep since they stem from the conscious mind and aren't viewed as spinal reflexes. Therefore, we envision our radio device assisting clinicians in itch evaluations during the differential diagnosis of such psychiatric conditions.

Our approach also holds benefits for drug development. Today, clinical trials rely on self-assessment tools like the NRS. These measures are subjective and lack sensitivity, leading to imprecise results and an amplified placebo effect \cite{Lee_2020}. To compensate for this variability, pharmaceutical companies often need to increase the number of participants in the trial, leadig to higher costs and longer trial duration. Notably, several pharmaceutical companies have approached us and are currently employing our radio device to monitor scratching and sleep in clinical trials \cite{emerald_sanofi}.

We also recognize our study's limitations. Notably, our study sample comprises only 20 chronic itch patients. While a broader participant base is ideal, our cohort aligns with the typical sizes of itch clinical trials \cite{trial_1, trial_2, trial_3, trial_4} and prior scratching detection studies \cite{Chun_2021, Yang_2023, Moreau_2018, Mahadevan_2021, Ji_2023, Ikoma_2019}. Expanding the study introduces challenges due to the significant manual effort required to analyze infrared (IR) camera footage. Nonetheless, our research represents one of the most comprehensive study in the context of scratching, encompassing 364 nights across multiple chronic itch conditions. Importantly, due to the vast data volume from our participants, all results in our study show statistical significance. Another limitation is that the AI system has only been tested for individuals who sleep without a bed partner. This, however, is a standard protocol when devising a new sensing system. For example, earlier works monitoring sleep and breathing \cite{Adib_2015c} via radio signals focused solely on solo sleepers. Subsequent studies broadened the technology to accommodate individuals with bed partners \cite{Yue_2018}. In a similar vein, we foresee future endeavors extending our scratching assessment model to settings with bed partners.

Despite its prevalence and debilitating impact, the field has lacked objective and sensitive methods for accurately assessing itch. Our study suggests that capturing the consequences of itch in humans can be done passively with zero overhead to the patient. The resulting measurements may lend new insight into our understanding of not only treatments for itch, but fundamental aspects of unconscious reflexes and their relationship to overall health.

\section{Materials and Methods}

\subsection{Study Design}
We conducted a 4-week observational research study, in which we enrolled 20 adults who were experiencing a range of chronic itch conditions—specifically, 4 cases of AD, 2 cases of PN, and 14 cases of CPUO. Despite the distinct characteristics of these conditions, they are uniformly evaluated through a shared endpoint (NRS). Therefore, our goal is to showcase the efficacy of our scratching evaluation across this diverse spectrum of conditions.
We carried out a 4-week observational study involving 20 adults with various chronic itch conditions: specifically, 4 cases of AD, 2 of PN, and 14 of CPUO. Even though these conditions have distinct features, they are all assessed using the same endpoint, NRS. Our objective was to demonstrate the effectiveness of our scratching assessment across this array of conditions.
In partnership with the dermatology clinic at Washington University School of Medicine and with the approval of IRB 201811086, we recruited and obtained informed consent from participants. We started with a baseline assessment, evaluating the severity of participants’ itch and reviewing their medication regimens. Those who qualified based on the protocol had our radio devices set up in their bedrooms, accompanied by an infrared video camera. We offered guidance on operating the camera, ensuring participants could switch it on and off to gather infrared video recordings. We asked them to activate the camera for a minimum of 4 nights weekly, while the radio device recorded data every night. Daily, participants provided their NRS scores. At the conclusion of the monitoring phase, we conducted a follow-up assessment, utilizing the same tools and questionnaires from the initial baseline evaluation.

\subsection{Ground Truth Labeling}
The video recordings were anonymized by blurring the faces of participants. These anonymized videos were then reviewed by human observers who annotated each instance of scratching. The task of annotating scratching from videos is intricate because many individuals sleep covered by blankets and frequently scratch beneath them. To enhance the accuracy of the annotations, each video was assessed on average by three separate labelers. A majority vote was taken to finalize the scratching labels. Labelers accessed the videos through a specialized labeling server using secure, password-protected accounts. They were presented with brief video segments, spanning several seconds, and were able to scroll through individual frames to mark the start and end of each scratching episode (as shown in fig. S1).

\subsection{Radio-Based Sensing Technology}
Recent advancements in radio-based sensing have paved the way for contactless monitoring of vital signs, gait, and sleep \cite{Kabelac_2019, Vahia_2020, Loring_2019, Adib_2013, Adib_2014, Adib_2015, Adib_2015b, Adib_2015c, Hsu_2017, Hsu_2017b, Li_2019, Tian_2018, Yue_2018, Zhao_2016, Zhao_2017, Zhao_2018, Zhao_2018_b, Kabelac_2020, Zeng_2020}. In this paper, we use the Emerald radio device, a tool frequently used in prior studies \cite{Kabelac_2019, Vahia_2020, Loring_2019, Adib_2013, Adib_2014, Adib_2015, Adib_2015b, Adib_2015c, Hsu_2017, Hsu_2017b, Li_2019, Tian_2018, Yue_2018, Zhao_2016, Zhao_2017, Zhao_2018, Zhao_2018_b, Kabelac_2020}, especially in previous research focused on sleep assessment via radio signals \cite{Hsu_2017b}. The device is mounted on the wall and consistently emits and receives low-power wireless signals that bounce off nearby people. Movements, such as scratching, modulate these radio signals. The device employs standard signal processing techniques like FMCW and antenna arrays to focus on radio reflections from the patient's sleeping area. These reflections are then processed by our AI model to detect instances of scratching.

\subsection{The AI Model}
We employ a neural network to detect scratching occurrences. The network uses reflected radio signals from the participant's sleeping area as input and classifies each moment into one of three categories: static, scratching, or motion (the latter encompasses all movements excluding scratching). The neural network is divided into three components: feature extractor, encoder, and decoder (as depicted in fig. S2).

Feature extractor: Given that scratching is a repetitive motion, our feature extractor aims to identify repetitive temporal patterns. This component accepts windows of radio signals, each lasting a minute, and processes them via multiple one-dimensional convolutional layers (specifically, 7 layers). This results in a sequence of feature vectors, which, when cross-correlated, yield a temporal similarity matrix (TSM). This matrix accentuates repetitive motions like scratching, as seen in fig. S3.

Encoder: The encoder takes as input the temporal similarity matrix generated by the feature extractor, and outputs a feature pyramid, which provides a high-level description of the input at different resolutions. The encoder is based on the EfficientNet \cite{efficientnet} architecture.

Decoder: The decoder takes as input the feature pyramid of the encoder, and outputs a time series of predictions that characterize the occurrence of scratching. The output of the decoder has the same length as the original radio signal (one minute). The decoder is based on the U-Net \cite{unet} architecture.

The model was implemented in PyTorch \cite{pytorch}, and trained using the cross-entropy loss and Adam optimizer. The weights of the model are randomly initialized at the start of training. The batch size is set to 32. The model was trained for 30K iterations on eight NVIDIA GeForce RTX 2080 Ti graphical processing units with distributed data parallelization.

\subsection{Model Interpretability}
We leverage the fact that scratching is a repetitive motion and design the first stage of the model to focus on temporal similarities in the input signals. As discussed previously, the first stage outputs a temporal similarity matrix (TSM). The TSM shows instances when the model discovers repetitive behavior, and hence, provides a form of interpretability for the model’s decisions.

Fig. S3 shows an example TSM. The horizontal axis of the TSM refers to time, and the vertical axis is a shorter (myopic) time axis that corresponds to how far ahead we look (relative to the position on the horizontal axis), to discover repetitive behaviors (i.e., scratching). The dark and yellow stripes in the TSM show a repetitive motion, and their separation reflects the repetition interval. In this example, at first, the participant was static as can be seen in the ground truth activity labels. Since a static scene leads to high correlation over time, being static shows as a bright yellow patch in the TSM, as shown in the left side of fig. S3a. Next, the participant moves their arm to start scratching. This short motion is highly uncorrelated with anything before or after, thus, it creates the darker triangular patch (right next to the initial yellow patch). Next, the participant starts scratching. Since, scratching is a repetitive motion, e.g., hand moving up and down, the TSM shows black and yellow stripes. The separation of the stripes indicates that scratching cycle is about 0.5 seconds. After scratching, the participant moves their arm back to a resting position. Again, this short motion creates a darker patch in the TSM (immediately after the first set of black and yellow stripes). At this point, the participant remains static for a little while (small, yellow triangle in the middle of the TSM), before they begin a second scratching bout.

\subsection{Repeated Measures Correlation}
We analyze the prediction accuracy of STH/SBH using the repeated measures correlation (rmcorr) \cite{Bakdash_2017}, which computes correlation while accounting for dependence between measurements from the same participant. The observed variable is the ground-truth video-based STH. The fixed effects are the predicted radio-based STH and the participant ID. The mathematical formula that describes the above linear model is as follows:

\begin{align*}
RadioSTH_{ij} &= \overline{RadioSTH_i} + ID_i\\
&+ \beta * (CameraSTH_{ij} - \overline{CameraSTH_i}) \\
&+ \epsilon_{ij}
\end{align*}

where $i$ is the participant index, $j$ is the per-participant night index, $\beta$ is the population-wide slope,  $\overline{CameraSTH_i}$ and  $\overline{RadioSTH_i}$ denote the per-participant mean value of the corresponding variable, and $\epsilon_{ij} \sim \mathcal{N}(0, \sigma_\epsilon)$ is the per-datapoint noise term.

A similar analysis is used to estimate the correlation between scratching and sleep metrics and NRS.

\subsection{Sensitivity and Specificity}
To evaluate the performance of scratching classification, we use sensitivity (recall), specificity, precision, receiver operating characteristic (ROC) curves, precision-recall curves (PR), area under the ROC curve (ROC AUC), and area under the PR curve (PR AUC). Sensitivity, specificity, and precision are calculated as (TP: True Positive; FN: False Negative; TN: True Negative; FP: False Positive):
\begin{align*}
\text{Sensitivity} &= \frac{\text{TP}}{\text{TP} + \text{FN}}\\
\text{Specificity} &= \frac{\text{TN}}{\text{TN} + \text{FP}}\\
\text{Precision} &= \frac{\text{TP}}{\text{TP} + \text{FP}}
\end{align*}
ROC curves demonstrate the trade-off between sensitivity and specificity as the classification threshold varies. PR curves demonstrate the trade-off between precision and recall as the classification threshold varies.  When reporting the sensitivity, specificity, and precision, we used a classification threshold of 0.5. We follow standard procedures to calculate the 95\% confidence interval for sensitivity, specificity, and precision. We also report the AUC, which is the area under the corresponding ROC or PR curves, showing an aggregate measure of detection performance.

\subsection{Software Packages}
All statistical analysis was performed with Python version 3.7 (Python Software Foundation) and R version 4.1 (R Foundation).

\section{Acknowledgements}

We are grateful to Kreshnik Hoti, Peter G. Mikhael, Raaz Dwivedi, Kelsey Auyeung, Blair Jenkins and Lydia Zamidar for their comments on our manuscript. We thank Malvika Joshi, Sarah F. Gurev, Shichao Yue and Rumen Hristov for their help with the labeling software, deployment and sleep model. We also thank all the individuals who participated in our study. M.O., M.Z., J.H., A.B., B.S.K. and D.K. are funded by NIH grant R01AR080392 and a grant from LEO Pharma. B.S.K. is also funded by National Institute of Arthritis and Musculoskeletal and Skin Diseases (NIAMS) grant R01AR070116.

\section{Author Contributions}
Conceptualization: D.K., M.Z., B.S.K. Methodology: M.O., M.Z., H.R., B.S.K., D.K. Data Curation: A.B., M.Z., M.O., J.H., H.R. Hardware: H.R., M.Z., D.K. Machine Learning Models: M.O., M.Z., D.K. Software (visualization and labeling): J.H., M.O., M.Z. Software (neural network and signal processing) M.O., M.Z., H.R. Validation and Formal Analysis: M.O., M.Z., D.K. Investigation: M.O., M.Z., H.R., B.S.K., D.K. Resources: B.S.K., D.K. Writing: M.O., M.Z., H.R., B.S.K., D.K Supervision: B.S.K., D.K. All authors reviewed and approved the manuscript.

\section{Competing Interests}
B.S.K has served as a consultant for AbbVie, Abrax, Amgen, AstraZeneca, Bristol Myers Squibb, Cara Therapeutics, Eli Lilly and Company, Escient Pharmaceuticals, Evommune, Galderma, GSK, Janssen Pharmaceuticals, LEO Pharma, Novartis, Pfizer, Recens Medical, Regeneron, Sanofi Genzyme, Trevi Therapeutics; he has stock in Locus Biosciences, and Recens Medical; he holds a patent for the use of JAK1 inhibitors for chronic pruritus; he has received grant, research, or clinical trial support from AbbVie, Cara Therapeutics, LEO Pharma, and Veradermics. M.Z. is an employee at Emerald Innovations and holds stocks in the company. D.K. and H.R. are cofounders of Emerald Innovations. D.K. receives research funding from the NIH, NSF, LEO Pharma, Sanofi, Takada, IBM, Gwangju Institute of Science and Technology, Michael J Fox Foundation, Helmsley Charitable Trust, the Rett Syndrome Research Trust, and serves on the scientific advisory board of Janssen and the data science advisory board of Amgen. The remaining authors declare no competing interests.

\bibliographystyle{unsrt}
\bibliography{main}

\end{document}



\title{Quantifying Itch and its Impact on Sleep Using Machine Learning and Radio Signals}

\author{{Michail Ouroutzoglou\textsuperscript{1}} \\
	\and
	{Mingmin Zhao} \\
	\and
	{Joshua Hellerstein} \\
	\and
	{Hariharan Rahul} \\
	\and
	{Asima Badic} \\
    \and
	{Brian S.~Kim\textsuperscript{1}} \\
	\and
	{Dina Katabi\textsuperscript{1}} \\
}
\date{}

\maketitle
\let\thefootnote\relax\footnotetext{\textsuperscript{1}To whom correspondence should be addressed. E-mail: michail@mit.edu, itchdoctor@mountsinai.org or dk@mit.edu}

\section*{Supplemental Material}

\setcounter{figure}{0}
\renewcommand{\figurename}{Fig.}
\renewcommand{\thefigure}{S\arabic{figure}}
\setcounter{table}{0}
\renewcommand{\tablename}{Table}
\renewcommand{\thetable}{S\arabic{table}}

\begin{figure*}[h]
\centering
\includegraphics[width=.8\linewidth]{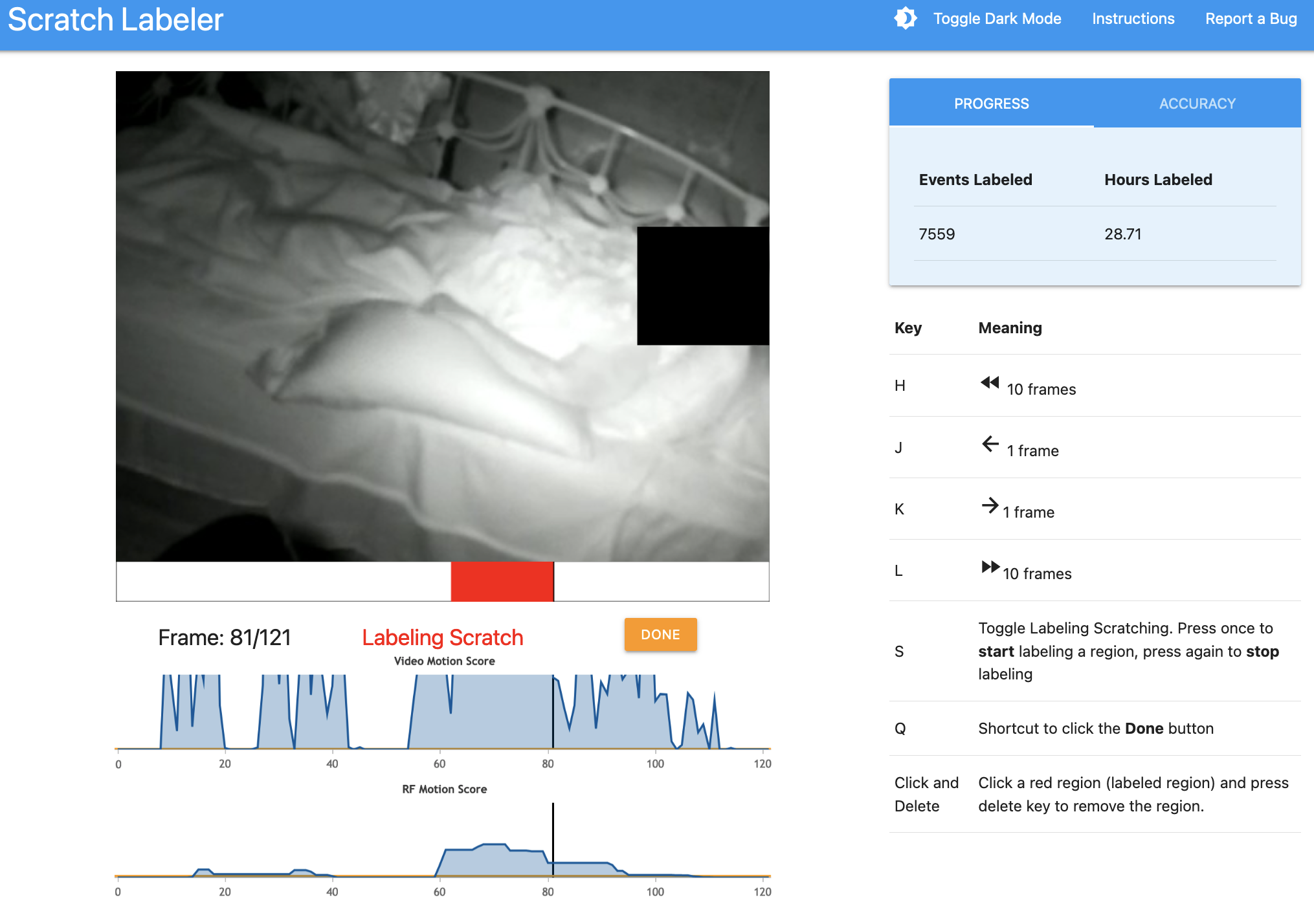}
\caption{\textbf{Labeling server.} A visualization of the labeling platform, where labelers were served short videos and asked to annotate the start and end of each scratching bout.}
\label{fig:labeling}
\end{figure*}

\clearpage 

\begin{figure*}[h]
\centering
\includegraphics[width=.8\linewidth]{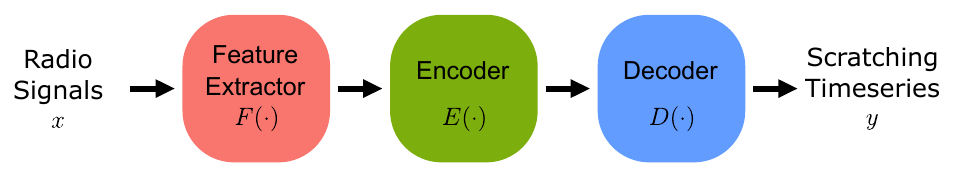}
\caption{\textbf{High-level overview of the neural network architecture.} The model consists of three stages: a feature extractor, an encoder, and a decoder. The model takes as input radio signals that are reflected from the participant’s bed location, and outputs a time series of scratching predictions.
}
\label{fig:model}
\end{figure*}

\begin{figure*}[h!]
\centering
\includegraphics[width=.8\linewidth]{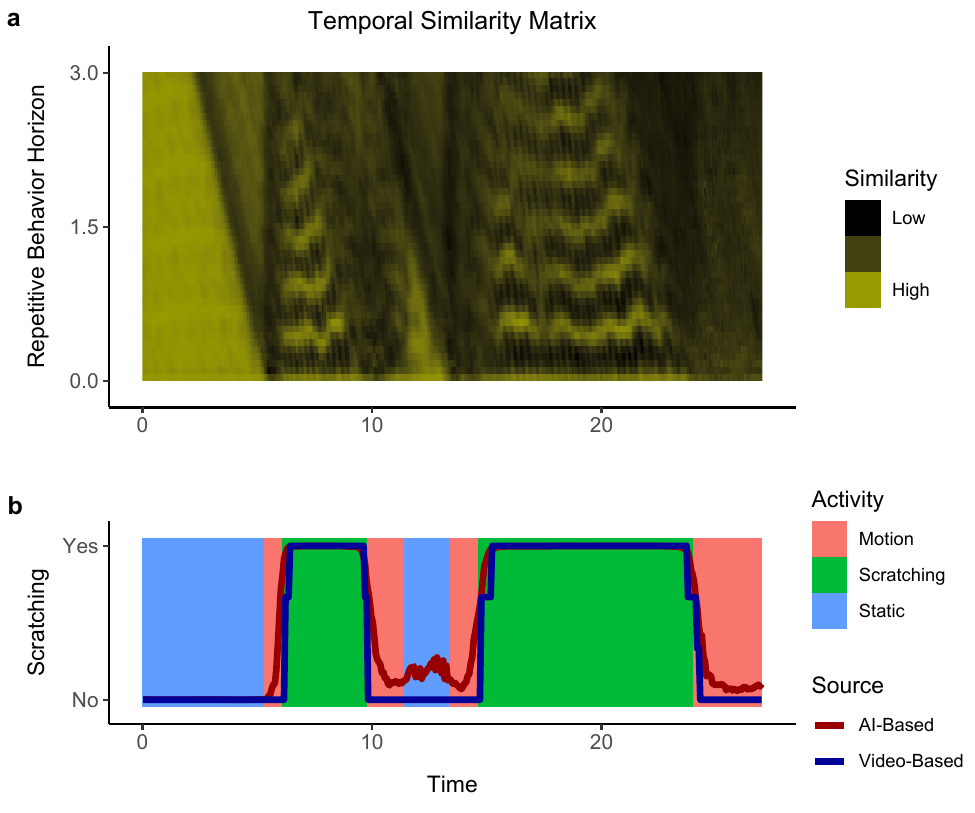}
\caption{\textbf{Scratching Visualization.} This figures visualizes a 26-second example during which the participant scratches twice. (a) shows the temporal similarity matrix, which highlights repetitive patterns (e.g., scratching). The horizontal axis refers to time. The repetitive yellow lines along the y axis refer to the repetitive movement of scratching. Bright yellow means high similarity, while black means low similarity. (b) The shaded regions show the ground truth where blue refers to static, green refers to scratching and pink refers to motion. The red curve shows model predictions, which aligns with the ground truth scratching.}
\label{fig:similarity}
\end{figure*}

\clearpage

\begin{table*}
\centering
\begingroup
\renewcommand{\arraystretch}{1.2}
\scalebox{0.85}{
  \begin{tabular*}{1.007\linewidth}{|p{0.09\linewidth}|p{0.1029\linewidth}|p{0.1029\linewidth}|p{0.1029\linewidth}|p{0.1029\linewidth}|p{0.1029\linewidth}|p{0.1029\linewidth}|p{0.1029\linewidth}|}
  \hline
  \cellcolor{gray!50}Work & \cellcolor{gray!50}Chun, et al. (2021) \cite{Chun_2021} & \cellcolor{gray!50}Yang, et al. (2023) \cite{Yang_2023} & \cellcolor{gray!50}Moreau, et al. (2018) \cite{Moreau_2018} & \cellcolor{gray!50}Mahadevan, et al. (2021) \cite{Mahadevan_2021} & \cellcolor{gray!50}Ji, et al. (2023) \cite{Ji_2023} & \cellcolor{gray!50}Ikoma, et al. (2019) \cite{Ikoma_2019} & \cellcolor{gray!50}Ours \\
  \hline
  \cellcolor{gray!50}Sensing Technology & \cellcolor{gray!10}Adhesive sensor on the back of the hand & \cellcolor{gray!10}Adhesive sensor on the back of the hand &\cellcolor{gray!10} Accelerometer devices on both wrists &\cellcolor{gray!10} Wrist-worn device on both wrists &\cellcolor{gray!10} AX6 actigraphy device on both wrists &\cellcolor{gray!10} Smartwatches on both wrists &\cellcolor{gray!10} Contactless radio sensor \\
  \hline
  \cellcolor{gray!50}Home Study &\cellcolor{gray!10} Y &\cellcolor{gray!10} Y &\cellcolor{gray!10} N &\cellcolor{gray!10} N &\cellcolor{gray!10} N &\cellcolor{gray!10} N &\cellcolor{gray!10} Y \\
  \hline
  \cellcolor{gray!50}Contactless &\cellcolor{gray!10} N &\cellcolor{gray!10} N &\cellcolor{gray!10} N &\cellcolor{gray!10} N &\cellcolor{gray!10} N &\cellcolor{gray!10} N &\cellcolor{gray!10} Y \\
  \hline
  \cellcolor{gray!50}Captures scratching by any body part &\cellcolor{gray!10} N &\cellcolor{gray!10} N &\cellcolor{gray!10} N &\cellcolor{gray!10} N &\cellcolor{gray!10} N &\cellcolor{gray!10} N &\cellcolor{gray!10} Y \\
  \hline
  \cellcolor{gray!50}Number of participants &\cellcolor{gray!10} 11 (pediatric) &\cellcolor{gray!10} 11 (adults) &\cellcolor{gray!10} 24 adults (6 healthy, 18 chronic itch) &\cellcolor{gray!10} 33 (adults) &\cellcolor{gray!10} 20 (adults) &\cellcolor{gray!10} 5 (adults)$^\dagger$ &\cellcolor{gray!10} 20 (adults) \\
  \hline
  \cellcolor{gray!50}Number of nights &\cellcolor{gray!10} 46 &\cellcolor{gray!10} 73 &\cellcolor{gray!10} 24 &\cellcolor{gray!10} 66 &\cellcolor{gray!10} 96 &\cellcolor{gray!10} 5 &\cellcolor{gray!10} 364 \\
  \hline
  \cellcolor{gray!50}Performance on long scratching events &\cellcolor{gray!10} Assumption: Report scratching only for dominant hand, and scratching events $\geq$ 4.5s\newline\newline Specificity: 99.3\%\newline Sensitivity: 84.2\%\newline F1-score: 82.9\%\newline
 &\cellcolor{gray!10} Assumption: Report scratching only for dominant hand, and scratching events $\geq$ 4.5s \newline\newline Specificity: 100\%\newline Sensitivity: 93\%\newline F1-score: Not reported
 &\cellcolor{gray!10} Assumption: Report scratching for events $\geq$ 2s \newline\newline Specificity: Not reported \newline Sensitivity: 66\% \newline F1-score: 68\%
 &\cellcolor{gray!10} Assumption: Report scratching for events $\geq$ 3s \newline\newline Specificity: 80\%\newline Sensitivity: 61\%\newline F1-score: 66\%
 &\cellcolor{gray!10} Assumption: Report scratching for events $\geq$ 3s \newline\newline Specificity: 80\%\newline Sensitivity: 64\%\newline F1-score: 44\%
 &\cellcolor{gray!10} Assumption: Report scratching for events $\geq$ 3s, excluding finger scratching \newline\newline Specificity: Not reported \newline Sensitivity: 90.2\%\newline F1-score: Not reported
 &\cellcolor{gray!10} Assumption: Report scratching for events $\geq$ 3s \newline\newline Specificity: 99.7\%\newline Sensitivity: 82.5\%\newline F1-score: 79.5\%
 \\
  \hline
  \cellcolor{gray!50}Performance on all scratching events &\cellcolor{gray!10} Not reported &\cellcolor{gray!10} Not reported &\cellcolor{gray!10} Not reported &\cellcolor{gray!10} Not reported &\cellcolor{gray!10} Not reported &\cellcolor{gray!10} Not reported &\cellcolor{gray!10} Specificity: 99.5\%\newline Sensitivity: 80.6\%\newline  F1-score: 75.3\% \\
  \hline
  \cellcolor{gray!50}Correlation of predicted scratching duration with ground truth &\cellcolor{gray!10} Not reported &\cellcolor{gray!10} Not reported &\cellcolor{gray!10} R = 0.945, but no significance reported &\cellcolor{gray!10} R = 0.82, p < 0.001 &\cellcolor{gray!10} Not reported &\cellcolor{gray!10} R = 0.901, p < 0.001 &\cellcolor{gray!10} R = 0.97, p < 0.001 \\
  \hline
  \cellcolor{gray!50}Performance on n=1 &\cellcolor{gray!10} Not reported &\cellcolor{gray!10} Not reported &\cellcolor{gray!10} Sensitivity, specificity, etc, are provided, as well as standard deviation &\cellcolor{gray!10} Not reported &\cellcolor{gray!10} Sensitivity, specificity, etc, are reported but no confidence intervals are provided &\cellcolor{gray!10} Not reported &\cellcolor{gray!10} R=0.71-0.99, p < 0.003 \\
  \hline
  \cellcolor{gray!50}Correlation with PROs &\cellcolor{gray!10} Not reported &\cellcolor{gray!10} Not reported &\cellcolor{gray!10} Not reported &\cellcolor{gray!10} Not reported &\cellcolor{gray!10} R = 0.44, but no significance reported &\cellcolor{gray!10} R = 0.21 p = 0.35$^\dagger$ &\cellcolor{gray!10} R = 0.22, p < 0.001 \\
  \hline
  \end{tabular*}
  }
  \caption{Comparison between different technologies for measuring nocturnal scratching. \textmd{$\dagger$: Sample size reported for studies with ground truth camera annotations.}}
  \label{table:2}
\endgroup
\end{table*}

\newpage
\onecolumn
\bibliographystyle{unsrt}
\bibliography{supplement}